\documentclass{article}

\usepackage{array}
\usepackage{arxiv}
\usepackage{pdfpages}
\usepackage[utf8]{inputenc} 
\usepackage[T1]{fontenc}    
\usepackage{hyperref}       
\usepackage{url}            
\usepackage{booktabs}       
\usepackage{amsfonts}       
\usepackage{nicefrac}       
\usepackage{microtype}      
\usepackage{lipsum}
\usepackage{subfigure}
\usepackage{graphicx}
\usepackage[section]{placeins}
\graphicspath{ {./images/} }

\title{Genetic Constrained Graph Variational Autoencoder (GCGVAE) for COVID-19 Drug Discovery}

\author{
 Tianyue Cheng \\
  University of California, Berkeley\\
  No. 2 Zongmao Toutiao, Xitiejiang Alley, Beijing \\
   \And
 Tianchi Fan \\
  University of Chicago\\
  No. 2 Zongmao Toutiao, Xitiejiang Alley, Beijing \\
  \And
 Landi Wang \\
  University of Notre Dame\\
  No. 2 Zongmao Toutiao, Xitiejiang Alley, Beijing \\
}

\begin{document}
\maketitle
\begin{abstract}
In the past several months, COVID-19 has spread over the globe and caused severe damage to the people and the society. In the context of this severe situation, an effective drug discovery method to generate potential drugs is extremely meaningful. In this paper, we provide a methodology of discovering potential drugs for the treatment of Severe Acute Respiratory Syndrome Corona-Virus 2 (commonly known as SARS-CoV-2). We proposed a new model called Genetic Constrained Graph Variational Autoencoder (GCGVAE) to solve this problem. We trained our model based on the data of various viruses' protein structure, including that of the SARS, HIV, Hep3, and MERS, and used it to generate possible drugs for SARS-CoV-2. Several optimization algorithms, including valency masking and genetic algorithm, are deployed to fine tune our model. According to the simulation, our generated molecules have great effectiveness in inhibiting SARS-CoV-2. We quantitatively calculated the scores of our generated molecules and compared it with the scores of existing drugs, and the result shows our generated molecules scores much better than those existing drugs. Moreover, our model can be also applied to generate effective drugs for treating other viruses given their protein structure, which could be used to generate drugs for future viruses. 
\end{abstract}

\qquad
\qquad 



\section{Introduction}
COVID-19 is a disease first diagnosed in Wuhan, China. The virus that causes COVID-19 is designated Severe Acute Respiratory Syndrome Coronavirus 2 (SARS-CoV-2) \cite{covid-19}. The most common symptoms of COVID-19 are fever, tiredness and dry cough \cite{covid-19_symptom,covid-19_symptom2}. With the progress of the illness, some patients experience aches and pains, nasal congestions, runny nose or sore throat. The age of patients has an important influence on the effect of this virus and most young people could recover from COVID-19. However, people aged over 60 are more susceptible to reaching severe conditions because of the virus. According to data from the WHO (World Health Organization), 20\% of COVID-19 patients suffer serious problems in breathing and need hospital care. With the quick spread of COVID-19 around the world, policies have been carried out to curb its effects, yet many are still exposed to severe health threats. Specifically, old people often have underlying medical conditions such as diabetes, heart diseases, respiratory disease or hypertension, which put them at greater risk if they are infected by the virus. Much effort has been put in developing vaccines for the virus. Though this is an effective solution to this situation, it takes a long time to develop, and is limited to a few countries with the best biochemical equipment. Computer-based drug discovery, on the other hand, is more accessible to developing countries. Therefore, our 
model has broad applications outside the COVID-19 context, namely it could be of use in similar scenarios for designing drugs for other diseases in the future. 

When coming to sorting out drugs based on existing databases and generate effective drugs, Machine Learning (ML) performs much more effectively and efﬁciently than humans \cite{covid-survey}. Since there are millions of ligands and protein of drugs in the National Center for Biotechnology Information (NCBI) database, a ML algorithm is certainly needed to efficiently sort out all useful ligands and protein and generate effective molecules. In this paper, our approach generates various useful molecules that could be used to treat COVID-19. We developed a generative model named Genetic Constrained Graph Variational Autoencoder (GCGVAE), which automatically sorts all useful data from the database and generates useful molecules to bind with SARS-CoV-2 protease. The performance (measured by binding afﬁnity, mentioned in section 6) of our molecules is much better than that of the existing drugs, including Remdesivir, Ribavirin, Umifenovir, Favipiravir, Lopinavir, etc. \cite{remdesivir,ribavirin,other_drug}. Those molecules have great potential to treat patients.

\begin{figure}[htbp]
\centering 
\includegraphics[width=0.95\textwidth]{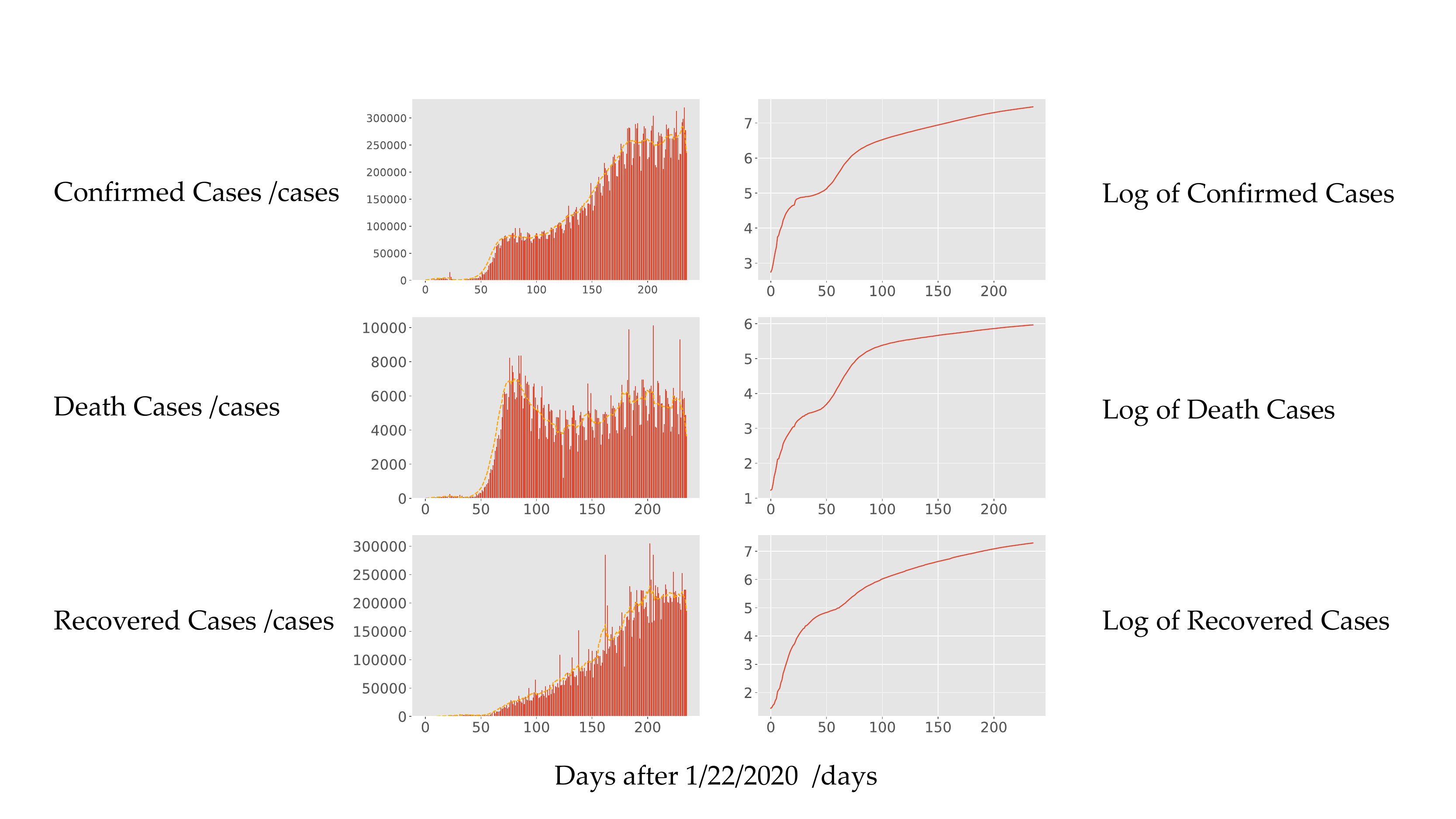} 
\caption{Global Epidemiological Statistics of COVID-19} 
\label{Fig.main2} 
\end{figure}

The figure above shows the statistics of the number of confirmed cases, death cases and recovered cases with its log value with respect to date (From 1/22/2020). The figure shows that the trend of the epidemic has caused severe damage to the society.
\section{Challenges}
In implementing GCEVAE, a batch of all active molecules must be selected for use in the first-generation sample that is to be fed into the model since using inactive molecules can reduce the effectiveness of the model and cause poorer performance. To sort out all active molecules from the original NCBI dataset, we first trained an edge-memory neural network (EMNN) \cite{emnn} for selecting usable molecule that is relatively active. The EMNN selects molecules based on the PubChem Standard Value \cite{pubchem}, a standardized value for indicating how active is a molecule, which corresponds to how much the molecule compound inhibits the given protein at a given concentration. 

In addition, calculating the composite score of the molecule (Introduced in section 6, Model Evaluation), is time-consuming. To shorten the training and inference time of the model, it only focuses on a specific atom in the molecule (Explained in section 4, GCGVAE model) instead of focusing on the entire molecule. Moreover, we use the software Autodock Vina \cite{autodock_vina} to calculate the binding affinity of a given molecule with the designated protein. Autodock vina embeds various optimizers that significantly shorten training time (Described in section 6, Model Evaluation).
 
We faced various challenges while tackling with the optimization model, namely the graph-based genetic algorithm (GB-GA). A chromosome in GB-GA is the encoded graph of molecules. Since the length of chromosomes is not ﬁxed but can vary to a great extent, we have to find a way to encode the molecules into chromosomes on which genetic operators can be applied. The traditional GA utilizes sample of ﬁxed lengths, which is relatively easy to implement genetic operators. While it is possible to ﬁnd a theoretical maximum molecule size and encode all molecules to have the same chromosome length as the maximum one, doing so greatly increases computational complexity. To deal with a generative problem in structural topology, I.Y. Kim et al \cite{GA_main2}, presented a method of starting with short chromosomes and increasing the length of chromosomes with the progression of generations.
However, since our GA is not for generating new molecules but for diversifying the results of our generative model, such a solution cannot be directly applied in this scenario. S. N. Pawar et al \cite{GA_main} presented another varied length GA for network intrusion detection, in which the length of chromosomes is independent of the number of generations. Moreover, the nonlinear structures of molecules (most medicinal molecules have ring structures, branches, etc.) brings additional challenges to our algorithms. In most of linear representations of molecules, the beginning and ending of each branch is denoted at their exact locations. In other words, if a branch originates after the third atom, the denotations will change and errors can be easily caused by confusion of the denotations while doing crossover and mutation. We solved the problem by placing all paired denotations of the molecules in the beginning when encoding them.
\section{Related Work}
\subsection{VAE}
Variational Autoencoder is a common generative model that uses one set of parameters to model the relationship between input and latent variables. It could be viewed as two coupled models that support each other. It was inspired by the Helmholtz machine (Dayan et al., 1995) \cite{vae_intro}. The framework of VAEs provides a computationally efficient way for optimizing DLVMs (deep latent-variable models) jointly with a corresponding inference model using Scholastic Gradient Descend (SGD). We optimize the variational parameters $\phi$ such that:
\begin{equation}
 q_\phi(z|x) \approx p_\theta(z|x) \label{XX}
\end{equation}
Here $q_\phi(z|x)$ is a parametric inference model, namely the encoder of VAE.$p_\theta(z|x)$ is the probability distribution function given by the model. The inference model can be any directed graphical model:
\begin{equation}
q_\phi(z|x) = q_\phi(z_1, \cdots, z_m|x) = \prod_{j=1}^M q_\phi(z_j|Pa(z_j),x)
\end{equation}
$Pa(z_j)$ is the set of parent variables of $z_j$ in the directed graph. The distribution of $q_\phi(z|x)$ can be parameterized using deep neural network.

The decoder part that learns to generate an output which belongs to the real data distribution is given a latent variable z as an input. This part maps a sampled z (initially from a normal distribution) into a more complex latent space (the one actually representing our data) and from this complex latent variable z, it generates a data point which is as close as possible to a real data point from our distribution. \cite{vae_bayes}

\subsection{Genetic Algorithm}
A genetic algorithm mainly follows Darwin's rule of nature and natural selection. It is generally consisted of a fitness calculator, a mutation function, a crossover function and a selection process; however, our genetic algorithm (see section 5.2) differs from the traditional genetic model in many ways.

The fitness function calculate scores of the population in the entire genetic algorithm. The selection algorithm filters the  population based on rules set beforehand. The mutation algorithm changes some parts of some chromosomes to increase result variability. The crossover process randomly pick samples and swap parts of chromosomes.

\subsection{AI Drug Development}
AI aided drug development mainly refers to generating small-molecule drug. Small-molecule drug discovery mainly focuses on chemically synthesized small molecules of active substances, which can be made into small-molecule drugs through chemical reactions between different organic and inorganic compounds. One group of AI-based drug development focuses on the discovery of new drug-like compounds at the molecular level. In \cite{2}, Beck \textit{et al.} proposed a DL-based drug-target interaction model (MT-DTI) to predict potential drug candidates for COVID-19. The authors collected the amino-acid sequences of 3C-like proteases and related antiviral drugs and drug targets from the databases of NCBI, Drug Target Common(DTC) \cite{4} and BindingDB \cite{5}.They also used a molecular docking tool called AutoDock Vina \cite{autodock} to predict the binding affinity between 3,410 drugs and SARS-CoV-2 3CLpro. Moskal et al. used AI to analyze the comparability of COVID-19 medicine and drugs involving similar indications to screen out second-generation drugs. They used the software Mol2Vec to convert molecular formulas into multidimensional vectors. In other words, each molecular formula is converted into a sentence of information, with each part of it being a word. Next, they used VAE to generate molecules that have suppresive qualities on the 2019-n-CoV virus. Then, they used CNN, LSTM, and MLP models to generate possibly funtional molecules, using 4456 drug-like molecules as training.

\subsection{Graph Neural Network}
GNN is a neural network inspired by CNN, which involves the use of complex multidimensional matrices to account for discrete characteristics. The core of CNN is local connections – sharing weights and using multi-layer calculations. Traditional CNNs can only deal with one or two dimensional problems, usually involving images and text; and GNNs are able to break these limits. The fact that GNN is a locally connected structure means that it can also use shared weights to reduce computational complexity \cite{dl}, making it an extensively applicable neural network. Multi-layer calculations is also the answer to many problems involving hierarchical pattern. 

The GNN model learns a state embedding \cite{sgt}$h_v \in \mathbb{R}$, which contains the information of neighboring nodes. State embedding $h_v$ is an s-dimension vector for $v$, which ultimately produces the result $O_v$. Let $f$ be a parametric equation, shared by each node, through which each node updates new information. Let $g$ be the local output function which describes how the output is produced. Thus $h$ and $O_v$ follow the following equation:
\begin{equation}
h_v = f(x_v,x_{co[v]},h_{ne[v]},x_{ne[v]})
\end{equation}
\begin{equation}
o_v = g(h_v,x_v)
\end{equation}
Here $x_v,x_{co[v]},h_{ne[v]},x_{ne[v]}$ are all features of v, respectively, edge, state, and features of the node and neighboring nodes.\\
Let $H,O,X$ and $X_n$ represent higher-dimension vectors that are the results stacking states. All outputs, features, and node features will be:
\begin{equation}
H = F(H,X) 
\end{equation}
\begin{equation}
O = G(H,X_N)
\end{equation}
Here, $F$ is the global transition function, and $G$ is the global output function. $F$ and $G$ stacks the $f$ and $g$ of each node.

The loss of the GNN, based on our current GNN structure and with the target information tv as supervision, is:
\begin{equation}
loss = \sum^{p}_{i=1}(t_i-o_i)
\end{equation}
Here, $p$ is the number of nodes in supervised learning, which minimizes loss by using gradient descent.

\section{GCGVAE Model}
Our generative model, GCGVAE, is mainly consisted of two segments: Constrained Graph Variational Autoencoder (CGVAE) and Graph-Based Genetic Algorithm (GB-GA). CGVAE is used to extract effective molecules from the ZINC dataset and generate drug molecules specifically for SARS-CoV-2 through altering the initial molecules in the ZINC dataset. CGVAE uses the process above to generate 1000 samples for the GB-GA module to begin process with, which would then perform SARS-CoV-2-specific alternations to the molecule samples in the input of GB-GA.

The whole process uses $N$ number of vectors as seeds that form a latent “specification” to let our model to generate new graphs ($N$ is the upper limit of the number of nodes generated). Then, by employing two decision functions, namely the focus function and the expand function, our model generates new molecules by setting up new edges. The focus function is used to choose nodes to connect and the expand function is used to choose edges to connect nodes that are chosen by the focus function. In a breath-first traversal, the focus function is implemented using deterministic queue (dequeue) with random choice for the inital nodes. 

The major point of our study is to let the expand function learn through training and thus generate an edge that can connect nodes that are selected by the focus function. To prevent an excessively large occupancy caused by massive amount of graphs learnt by the expand function \cite{graph_h}, our expand function trains through partial graph structure. 
\begin{figure}[htbp]
\centering 
\includegraphics[width=0.7\textwidth]{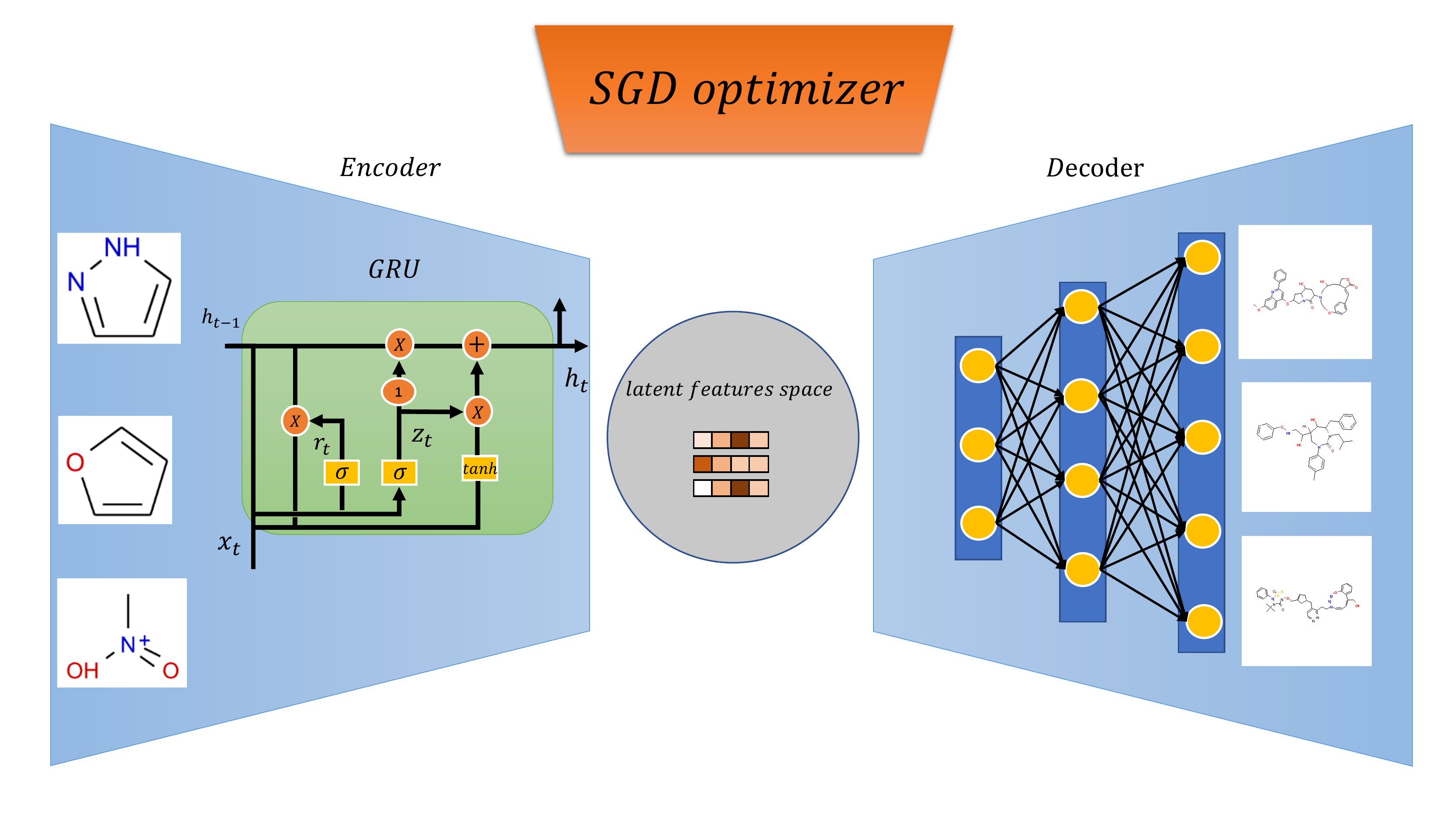} 
\caption{The scheme of CGVAE model. \quad GRU:Gated Recurrent Units \quad SGD: Scholastic Gradient Descend}
\label{Fig.main2} 
\end{figure}

The figure below shows the general structure of our CGVAE model. The encoder is consisted of a GRU unit that forms the GGNN network. The decoder is a normal VAE decoder. We optimize the model by using Scholastic Gradient Descend.

\paragraph{Initialize nodes}
We firstly set up state $h_v$ and every node $v$ in an initial graph that is not connected. Specifically, We assume the initial value of the graph structure $Z_v$, which is drawn from a standard normal $\mathcal{N}(0,I)$ in $d$ dimension. $h_v$ is the integral form of $[Z_v,T_v]$. Here $T_v$ is an explainable interpreter that is used to indicate the type of nodes. $T_v$ is the mapping of $\tau_v \sim f(z_v)$ that is obtained through $Z_v$ by learning from the output of the model, where $f$ represents the whole neural network. The explainable part of $h_v$ gives us a method to set limits with force. \\
According to these variables, we are able to calculate global representation $\mathbf{H}_t$ (the distribution and constitutions of every node at step $t$) and $\mathbf{H}_{init} = \mathbf{H}_0$ (the distribution and constitutions of every node at step 0). At the same time, we also set a stop node to terminate the training. 

\paragraph{Update nodes}
Whenever we obtain a new graph , we give up the original $h_v$, calculate a new $h_v$, and update for all nodes (achieved by updating information of neighboring nodes). Here we employed the standard gated graph neural network (GGNN) $G_{dec}$ for $S$ steps. 
\begin{equation}
m_v^{(0)} = h_v^{(0)} \qquad m_v^{(s+1)} = \mathbf{GRU}[m_v^{s},\sum_{v\stackrel{l}{\leftrightarrow}u}{E_l(m_u^{(s)})] \qquad h_v^{(t+1)} = m_v^{(S)}} 
\end{equation}
$E_l$ here is an edge-type specific neural network \cite{emnn}. $GRU$ here mainly describe how the GGNN works \cite{ggnn}.

\paragraph{Selecting edge and labeling}
First we choose a node $v$ from the queue. Expand function choose edges from $v$ to $u$: $v\stackrel{l}{\leftrightarrow}u$. The length of the label is l. For other nodes that are not focuses, we build a feature vector $\phi_{v,u}^{(t)} = [h_v^{(t)},h_u^{(t)},d_{v,u},\mathbf{H}_{init},\mathbf{H}^{(t)}]$. Here, $d_{v,u}$ is the graph distance between two nodes so that we can let the model obtain information on the nodes and edges. We would use the above information to generate specific distributions of edges. \\

\begin{equation}
p(v\stackrel{l}{\leftrightarrow}u | \phi_{v,u}^{(t)}) = p(l | \phi_{v,u}^{(t)},v{\leftrightarrow}u)\cdot p(v{\leftrightarrow}u | \phi_{v,u}^{(t)})
\end{equation}

\begin{equation}
p(v{\leftrightarrow}u | \phi_{v,u}^{(t)}) = \frac{M^{(t)}_{v{\leftrightarrow}u}\exp[C(\phi_{v,u}^{(t)})]}{\sum_w{M^{(t)}_{v{\leftrightarrow}w}}\exp[C(\phi_{v,w}^{(t)})])},\qquad p(l | \phi_{v,u}^{(t)}) = \frac{m^{(t)}_{v\stackrel{l}{\leftrightarrow}u} \exp[L_l(\phi_{v,u}^{(t)})]}{\sum_k{m^{(t)}_{v\stackrel{k}{\leftrightarrow}u}} \exp[L_k(\phi_{v,u}^{(t)})]}
\end{equation}

\section{Training}
The above model is dependent upon trained networks   (trained by a VAE architecture using the graph dataset D), as well as a latent space where the latent parameters of the model reside.
\subsection{CGVAE}
Our VAE is encoded by a GGNN $G_{enc}$, which is parametrised by the mean $\mu_v$ and the standard deviation $\sigma_v$ vectors. In a latent space with $d$ dimensions, we embed each node in an input graph $\mathcal{G}$ to a diagonal normal distribution, from which the latent vector samples $z_v$ are taken. The $\mathbf{KL}$ divergence between the encoder distribution and the standard Gaussian prior is measured by the usual VAE regulariser term:
\begin{equation}
\mathcal{L}_{latent} = \sum_{v \in \mathcal{G} }{\mathbf{KL}(\mathcal{N}( \mu_v ,diag(\sigma_v)^2)||\mathcal{N}(0,I))}\end{equation}

During training, each generation is conditioned by an inactive sample from the encoder distribution. Training of the overall model is supervised by generation traces obtained from graphs in $D$.
\\
\paragraph{Node Initialization}
Let node permutations be denoted by $\mathcal{P}$, and let $h_v(t=0)$ denote the initial node states. From each node $v$ a sample of the node specification $\mathbf{z}_v$ is taken, and using $f$ (the learned function), the label $\mathbf{\tau}_v$ is generated. The likelihood of the labels $\mathbf{\tau}^*_v$(observed in the encoded graph) being re-generated is \\
\begin{equation}
p(\mathcal{G}^{(0)}|\mathbf{z}) = \sum_\mathcal{P}{\mathbf{\tau} = \mathcal{P}(\mathbf{\tau}^*)|\mathbf{z}} > \prod_v{p(\tau_v = \tau_v^* | \mathbf{z}_v)}
\end{equation}
Since node type $\tau^*_v$, and therefore $Z_v$, is known, the single contribution from the encoder ordering gives a lower bound, which can undergo further improvement given a set2set model \cite{set2set}.
\paragraph{Edge Selection and Labelling During training}
The sequence of edge additions is supervised by basing the supervision upon breadth-oriented traversals of all graphs in $D$. A Monte-Carlo estimate, using a miniset of sampled traces, is taken of the marginal. This is done instead of computing the logged possibility, since doing so has the disadvantage of not being computationally tractable.\\
Let $\mathcal{G}^{(0)}$ represent the initial collection of unconnected nodes. From the function below, we can see that the lower bound of the log probability of a graph $\mathcal{G}$ is dependent upon the log probability of all traces.
\begin{equation}
\log p(\mathcal{G}|\mathcal{G}^{(0)}) = \log \sum_{\pi \in \Pi}{p(\pi | \mathcal{G}^{(0)})} \geq \log(|\Pi|) + \frac{1}{|\Pi|}\sum_{\pi \in \Pi}{\log p(\pi|\mathcal{G}^{(0)})}
\end{equation}
Let v denote the current focus node, and let $\epsilon = v\leftrightarrow u$ be the edge added at step $t$. Every trace of each generation $\pi \in \Pi$ can be split into a series of steps, which are ordered as $t, v, \epsilon$. 
\begin{equation}
\log p(\pi| \mathcal{G}^{(0)}) = \sum_{(t,v,\epsilon)\in \pi}\left\{{\log p(v|\pi,t)+\log p(\epsilon|G^{(t-1)},v)}\right\}
\end{equation}
The first term of this equation is the choice of focus node at step $t$ of trace $\pi$, which is a uniform distribution for the first focus node, and after which follows a breadth-first search. The set of generation states $S$, which includes the features of both the node we are currently focusing and the neighboring nodes around it, is put into consideration in order to elevate the model further.

Let $|\mathcal{E}_s|$ denote the multiplicity of state $s$ in $\Pi$ – the number of node traces containing graph $\mathcal{G}^{(t)}$. We use $\mathcal{E}_s$ to represent edges from focus node $v$ that are presented in graph $\mathcal{G}$. Both of the above are uniform, so we can rearrange and sum the value of them over steps.
\\
\begin{equation}
\frac{1}{|\Pi|}\sum_{\pi \in \Pi}\sum_{(t,v,\epsilon)\in\pi}{\log p(\epsilon|s)} = \frac{1}{|\Pi|}\sum_{s \in S}\sum_{\epsilon \in \mathcal{E}_s}{\frac{|s|}{|\mathcal{E}_s|} \log p(\epsilon|s)} = \mathbb{E}_{s \sim \Pi} [ \frac{1}{|\mathcal{E}|}\sum_{\epsilon \in \mathcal{E}_s}{\log p(\epsilon|s)}]]
\end{equation}
In this equation, $|s|/|\Pi|$ represents the probability of observing state s in a random draw from all states in $\Pi$. We define the loss fucntion of model as:
$$
\mathcal{L}_{recon} = \sum_{\mathcal{G} \in \mathcal{D}}{\log [p(\mathcal{G}|\mathcal{G}^{(0)})\cdot p(\mathcal{G}^{(0)}|\mathbf{z})]}
$$ 
Here again, a Monte Carlo estimate is taken from the set of numbered generation traces, which causes variance estimate to be high. 

Local optimization of the generated graphs is needed. This is done through gradient ascent (note: this is not gradient descent) in continuous latent space using a differentiable gated regression model. $g1$ and $g2$ are neural networks, and $\sigma$ is the sigmoid function.\\
The equation is as follows:
\begin{equation}
R(\mathbf{z}_v) = \sum_{v}{\sigma(g1(\mathbf{z}_v))\cdot g2(\mathbf{z}_v)}
\end{equation}
$L2$ distance is used to compute loss during training; and in testing, an initial latent point $\mathbf{z}_v$ can be sampled and used to gradient-ascent towards a local optimal point $\mathbf{z}_v^{*}$, which is kept within the standard normal. $Q$, the optimized property, is ultimately produced.

The main objective of our training is $\mathcal{L} = \mathcal{L}_{recon}. + \lambda_1\mathcal{L}_{latent} + \lambda_2\mathcal{L}_Q$, consisting of normal VAE objective. Note that VAE loss of Yeung et al.  \cite{vae_optimize} is not strictly followed.

\subsection{Graph-Based Genetic Algorithm}
Our genetic algorithm takes molecules from the CGVAE model as input, on which various genetic operators are applied, and produce desirable molecules as output.
\begin{figure}[htbp]
\centering 
\includegraphics[width=1.0\textwidth]{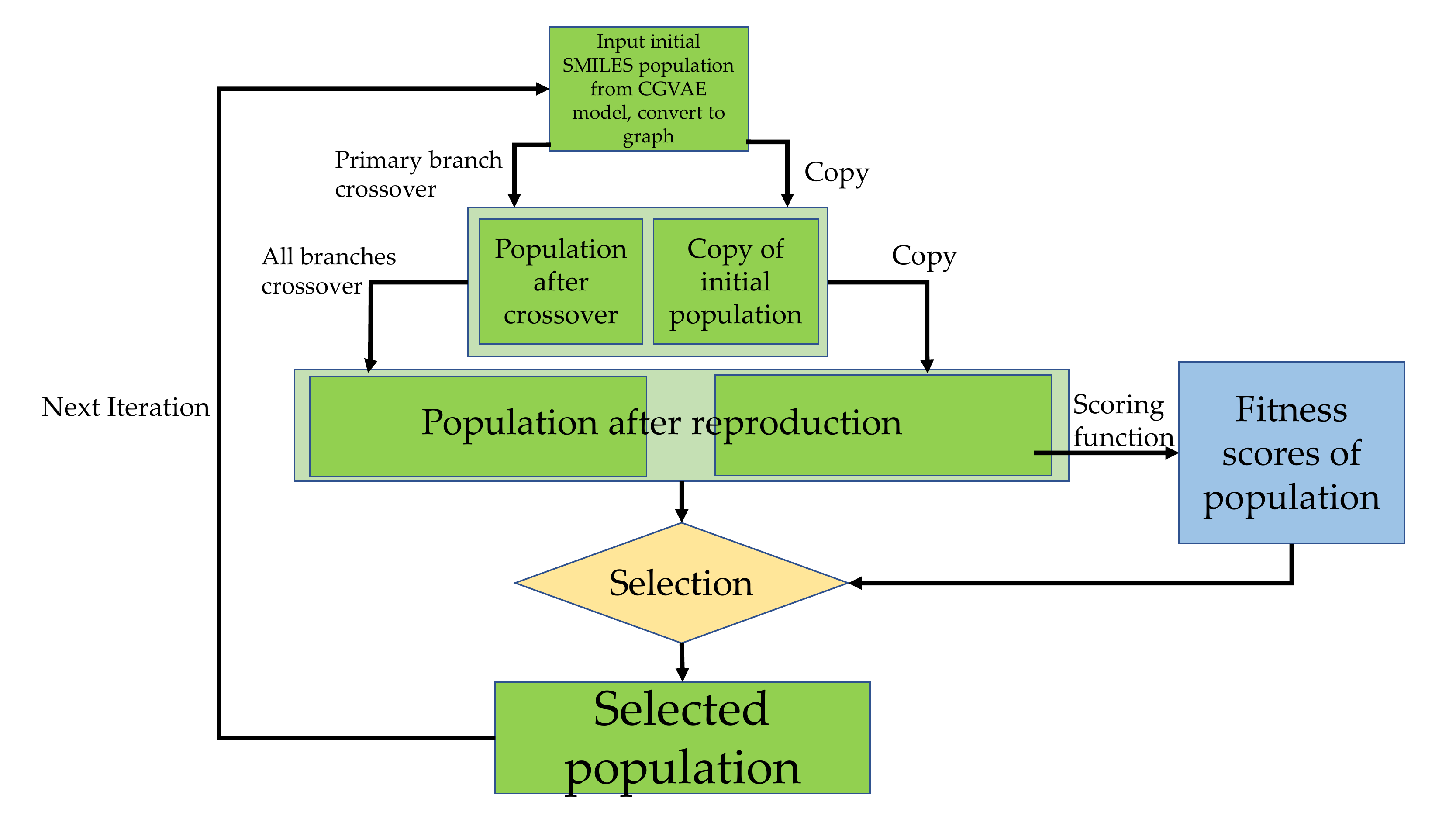} 
\caption{The schematic map of our Genetic Algorithm} 
\label{Fig.main2} 
\end{figure}

Figure 3 describes the process of our Genetic Algorithm. It is further explained in the following sections.

\paragraph{Population Initialization}
Since molecules share great similarities with undirected graphs, we use graphs to represent each molecule instead of using merely SMILES strings. Each node of the graph represents an atom of the molecule. And each edge represents the bond between atoms in the molecule. The graph structure is stored for further generation. 

\paragraph{Crossover \& Mutation \& Reproduction}
Our system of crossover, mutation, and reproduction is based on branch-switching, i.e. the switching of randomly chosen branches between two given molecules. Our reproduction process is embedded in the crossover function, and mutation is not directly implemented (instead, it is emulated by a small-scale crossover). Because crossing over molecular branches produce nearly unpredictable results, we create a copy of the original population each time we perform a crossover, and combine it with the crossed over population. This way, if our crossover creates undesirable results, successful individuals from the original population can still succeed into the next round. Note that in this process of crossover and copying, the population is doubled, saving the need for a sperate reproduction algorithm.

For each generation, we perform two rounds of crossover and copying. In the first round, only the upper-most branches of molecules are swapped. Since the upper-most branches of molecules have little variability (most often, they are singular carbon or oxygen atoms), swapping them creates only minor differences, if any. Therefore, the population created after this wave of crossover differs little from the copied one, much like a mutated population in the traditional GA. Thus, the mutation process is emulated by this round of crossover.

In the next round of crossover and copying, more significant changes are applied. Instead of uppermost branch switching, universal branch switching is performed. In other words, the choice range of switchable branches is changed from only uppermost branches to all branches. Note that after this round, the population is four times that of the original one; of this current population, one quarter is identical with the original one, one quarter has minor differences with the original one (this is the quarter crossed over in the first round and copied in the second), and half (the part of the population crossed over in the second round) has great differences with the original. Thus, the current population is a mixture of (with respect to the original one) no changes, minor changes, and significant changes.

\begin{figure}[htbp]
\centering 
\includegraphics[width=1.0\textwidth]{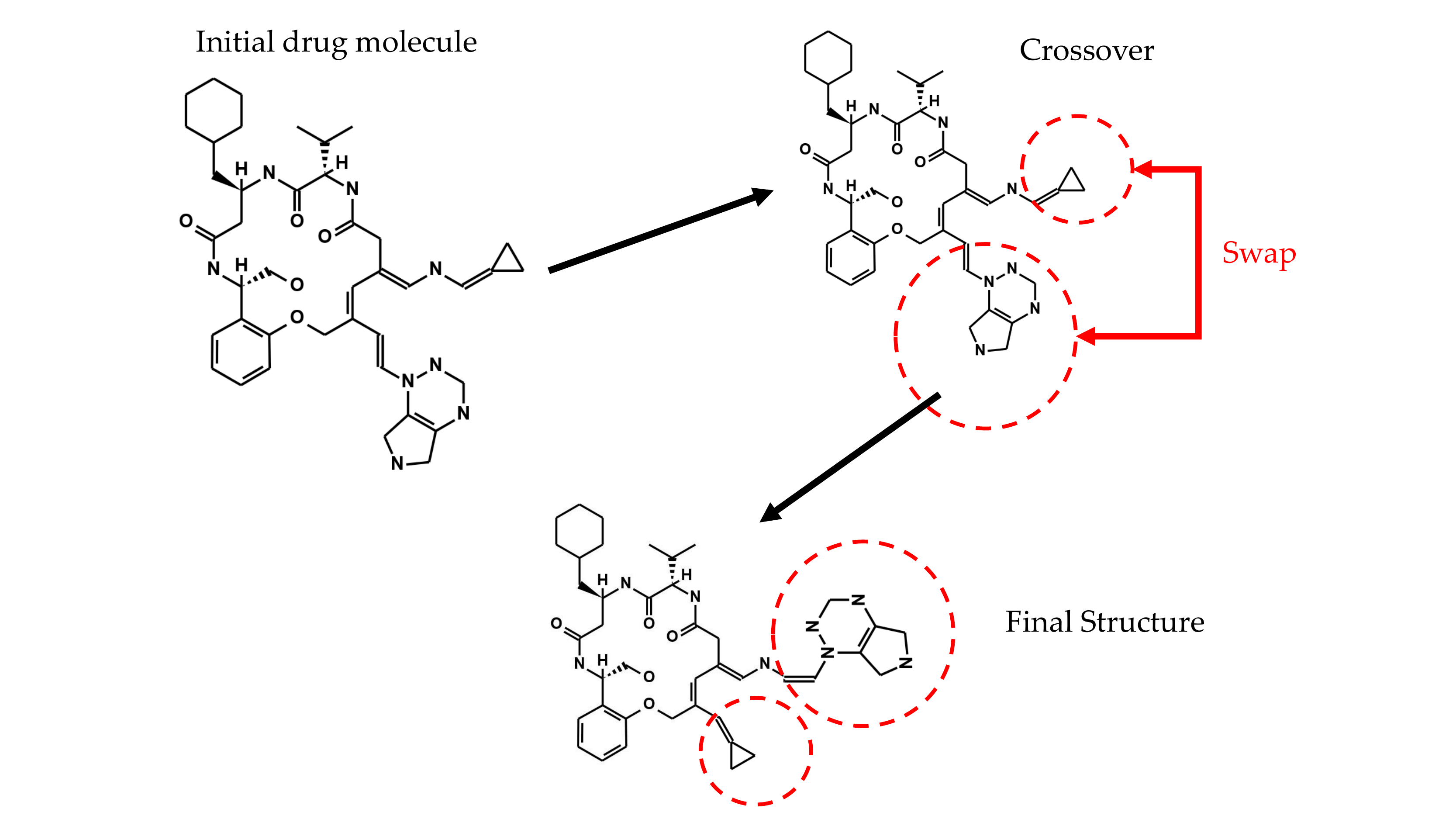} 
\caption{The process of crossover and mutation} 
\label{Fig.main2} 
\end{figure}

The figure above illustrates the process of mutation and crossover in our GA model. A molecule that is to be mutated is input into our mutation algorithm. When a crossover/mutation process takes place, two functional groups (circled in the figure) in one pair swap each other. It will generate a new molecules as a results.

\paragraph{Fitness Scoring}
We define the final score of a molecule by calculating a composite score due to different factors. The composite score is the sum of weighted score of the binding affinity of a molecule (Explained in section 6, model evaluation), the validity test function (checking the validity of a molecule) and a size penalty score (We give penalty to larger molecules since they are hard to be synthesized and unstable, explained in section 6). The scoring function returns the composite score for selecting population.

\subsubsection{Selection}
Our selection function uses a combination of Elite Retention Selection and Simple Random Selection. It is basically about individuals within the population with the highest fitness scores (the top 25\%) is automatically retained, and another 25\% is randomly selected from the remaining population. The Elite Retention Selection is implemented in order to retain good results that have been produced, and the Simple Random Selection ensures result variability (since changes to molecule structure produce largely unpredictable results, and low-scoring molecules may crossover into high-scoring ones).

\section{Model Evaluation}

\subsection{Molecule availability checking using valency masking}
Atomic valency is indicative of the number of molecular bonds an atom can make to form a stable molecule, and there are rules predicting it that can be used to construct structurally sound molecules. In our model, every type of node is given a fixed valency based on current chemical knowledge. For example, $\mathbf{Cl}$ (a chlorine atom) has valency 1. $M$ and $m$ are used to ensure that the number of bonds $b_v$ remain at a reasonable level and does not exceed $b_v^*$ \cite{rdkit}. This ensures that only valid molecules (which are parsable via RDKit \cite{rdkit}) are generated. Morover, $M^{(t)}_{ v \leftrightarrow u}$, which addresses this problem, is:
\begin{equation}
M_{v \leftrightarrow u}^{(t)} = \mathbf{I} (b_v < b_v^*) \times \mathbf{I} (b_u < b_u^*) \times \mathbf{I} (no\quad v \leftrightarrow u\quad exists) \times \mathbf{I} (v \neq u) \times \mathbf{I} (u\quad is\quad not\quad closed)
\end{equation}
$F$ is an indicator function. And $m^{(t)}_{v\stackrel{l}{\leftrightarrow}u} = M^{(t)}_{v\stackrel{l}{\leftrightarrow}u} \times \mathbf{I}(b_u^* - b_u \leq l)$ is defined using $l$ to represent the type of bond.
\subsection{Virtual Screening}
Virtual screening of drugs has become the standard technology in modern day drug discovery, which can be done through molecular docking. Molecular docking is a technology performed on computers in order to indicate how molecules bind with each other, in this case especially large protein receptors and small ligands. Molecular docking is realized by using the “lock and key” principle, which compares large protein molecules to locks and small ligands to keys. Each ligand fit into appropriate positions in protein molecules according to their special quantities such as conformation, orientation, and quantities, like keys fitting into locks. What molecular docking tries to achieve is in nature an optimization problem, seeking to find a best-fit of ligands into receptors. Molecular docking simulates the process of binding between receptors and ligands and show their mode of binding as well as their binding affinity, which can thus be used to do virtual screening of drugs. \\
\\
The specific molecular docking software used in this paper is Autodock, an open-source software that enables its users to execute ligand-protein docking effectively \cite{autodock}. Autodock is widely used in the field of computer science and has been cited for most times compared to other molecular docking software. The reason we use Autodock is that other scientists in the field are already using it to test effectiveness of drugs inhibiting SARS-CoV-2, which proves its feasibility on drug discoveries. \\
\\
The process of Autodock is as followed \cite{autodock_n}: First, a protein molecule is input into the docking software and waiting to be scanned. The purpose of scanning is to look for amino acid residues surrounding the active site of the protein (receptor). A box is then formed surrounding the protein with grid points and to be scanned by atom probes: atom probes are different types of atoms within the structure of ligands. The software would take the atom probes and scan the grid points within the box and calculate energy needed to bind with atoms in the receptor. The last step is that the software take the ligand to do a conformational search within the box and calculate a score correspondent to a specific ligand using a scoring function according to different conformations, orientations, positions, and energy \cite{f18,f18_2}:
\begin{equation}
\Delta G_{bind} = \Delta G_{solvent} + \Delta G_{conf} + \Delta G_{init} + \Delta G_{rot} + \Delta G_{t/t} + \Delta G_{vib}
\end{equation}
The components of the equation are solvent effects, conformational changes in the protein and ligand, free energy due to protein-ligand interactions, internal rotations, association energy of ligand and receptor to form a single complex and free energy due to changes in vibrational modes. A low (negative) energy indicates a stable system and thus a likely binding interaction. 

To increase the speed of evaluating effectiveness of drugs, we used Autodock Vina, Autodock’s successor. Autodock Vina has its advantages and limitations. Autodock Vina, compared to Autodock, has an improved local search routine and makes use of multi-CPU setups, which is the main cause for its increased speed. Autodock vina also utilizes several optimizing methods, including genetic algorithm, simulated annealing, and particle swarm optimization \cite{autodock_vina,autodock_vina2,autodock_vina3,autodock_vina4,autodock_vina5}. One major difference between Autodock and its successor is that distances between grid points within the box in Autodock can be defined to smaller sizes than 1 (e.g. 0.37), while the distance between grid points in Autodock Vina can only be set to 1, which might lead to leaving out possible best-fits. 
\\
\paragraph{Binding Affinity}
Binding affinity is a quantitative index used to measure the strength of molecular interactions, especially between a large protein molecule and a small ligand binding partner. The better binding between protein and ligand molecules the larger the binding affinity, which is measured by Kcal/mol. Binding affinity is influenced by non-covalent intermolecular interactions such as hydrogen bonding, hydrophobic, electrostatic interactions and Van der Waals forces between the molecules. In addition, binding affinity between a ligand and its target molecule may be affected by the presence of other molecules (which cannot be observed in our situation). \\
The measuring method of binding affinity is to measure the equilibrium dissociation constant $K_D$. Equilibrium dissociation constant is a specific type of equilibrium constant that measures the propensity of a larger object to separate (dissociate) reversibly into smaller components. The larger the equilibrium dissociation constant is, the smaller the absolute value of the binding affinity is, the easier the ligand is to separate from the large protein molecule, or the less likely the ligand is to bind with the large protein molecule; the smaller the equilibrium dissociation constant is, the larger the absolute value of the binding affinity is, the harder the ligand is to separate from the large protein molecule, or the more likely the ligand is to bind with the large protein molecule. For drug discovery, measuring binding affinity can help to rank order hits binding to the target and design drugs specifically for protein molecules.\\
By using Autodock Vina, we are able to calculate the binding affinity between the SARS-CoV-2 main protease (6LU7) and the drug ligands on computers and output the rank of molecules based on whether they are able to bind with the main protease. However, the binding affinity measured does not convey a specific or absolute value, but rather a comparative rank that only determines whether a molecule is better at binding with a protein molecule than another one. \\
However, there is still a method of comparing the drug molecules we generated to the molecules that already exist now and to molecules generated by other people. By retrieving the SMILES of drugs of other sources, we are able to convert them into $pdbqt$ format and do docking with the main protease of SARS-CoV-2 again on our device and compare with the drugs generated by our model.

The figure below is the docking pose of one generated ligand with 6LU7. The docking of ligands with protein receptors depends on properties of the ligand and the protein receptor, such as conformation, orientation, position, and energy. The graph below describes how the ligand bind with the protein In both pictures the protein (6LU7) is represented in blue.

The table below is the binding affinity of our generated molecules and existing drug molecules. Our molecules are named CFW. Others are those existing drugs' molecules. Our results shows significant improvement in effectiveness of inhibiting SARS-CoV-2019.
\begin{figure}[htbp]
\centering 
\includegraphics[width=1.0\textwidth]{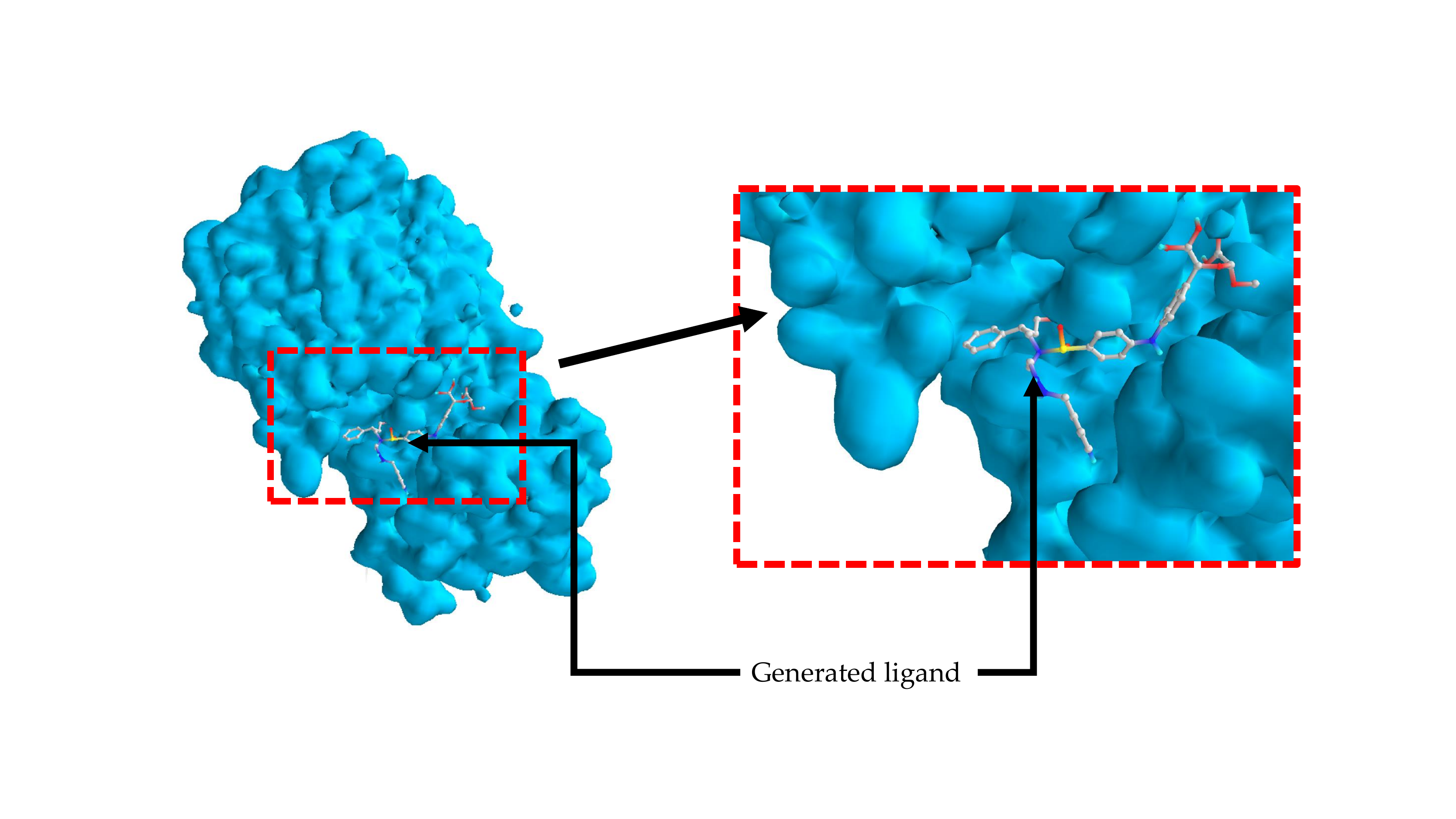} 
\caption{The Docking pose of our molecule with SARS-CoV-2019 protein} 
\label{Fig.main2} 
\end{figure}

\begin{center}
\begin{tabular}{|p{0.33\columnwidth}|p{0.33\columnwidth}|p{0.33\columnwidth}|}
\hline
Name of the drug& SMILES& Binding Affinity\\
\hline
CFW-001 & C[C@@H]([C@@H](CC/C=C($\backslash$CC)/c\newline1c(ccc2c1/N=C$\backslash$C(=C(/[C@@H]1[C@\newline@]34[C@H]([C@H]13)[C@@H]4/C/1\newline=C(/[C@\newline H](/C=C/c3c(/C=C/O1)cccc3)O)\newline$\backslash$[C@H]2O)$\backslash$O)$\backslash$O)O/C(=C/[C@](N)(\newline Cc1ccccc1)CO)/O)CO)OC.O.O
& -11.9\\
\hline
CFW-002 & C/C/1=C/CCc2cccc(c2NC(=O)N2[C\newline@@H](C)C=C[C@H]3CNN(/C=C(/C\newline4=C[S]=C(/C=C$\backslash$C1)O4)$\backslash$CC[C@]\newline(C)(F)C[C@]1(CCC(=O)C(=O)\newline O1)O)C=C23)O
& -9.9\\
\hline
CFW-003 & C=C1c2cccc(c2)N2N[C@@H](C3=CC\newline=CO[C@H]3[C@@H]3C(=O)N1C(=O)\newline OC[C@@H](CC)C(=O)O[C@@H]1C\newline[C@@H]1NC3=O)N[C@H]2C
& -9.7\\
\hline
CFW-004 & CC1=C[C@]2(C=C1)Nc1sc(cc1[C@@\newline H](C)/C=C/C)NC(=O)[C@H]1[C@@H\newline]([C@H]([C@H](N(C(=O)N1Cc1cc(c(c\newline c1C)O2)C1CC1)Cc1cccc(c1)C(=O)NO)\newline Cc1ccccc1)O)O
& -9.6\\
\hline
CFW-005& C$\backslash$1(=C/C=C($\backslash$C(=O)Oc2ccc(C)cc2O[C\newline@H]2C[C@H]3C(=O)N[C@@H]([C@\newline H](CNNC(=O)[C@@H]1[C@H](C(=O\newline)N3C2)C1=CC=CC1)O)Cc1ccc2N[C@\newline@H](N)Nc2c1)/O)/C=C
& -9.4\\
\hline
CFW-006& C/C/1=C$\backslash$C=C$\backslash$2/C[C@@H](C\newline CC[C@H]3C[C@@H]4CCOCc\newline 5ccc(c(c5)C(=O)N[C@H](Cc5ccc(\newline cc5)S(=O)(=O)N4)COC(=O)N[C@@\newline H](C1=O)C(C)(C)C)[C@H](NC3=O)\newline CO)NS(=O)(=O)N2
& -9.3\\
\hline
CFW-007 & [C@@H]1(C[C@H]2C[C@@H]3CC4\newline=C[C@H](C2)[S]13(NC[C@]12CN(C)\newline C3=C(C[C@@H](CCCc5ccc(c(F)c5)[\newline C@H](N1)OC(=O)C(=O)OC)[C@@H]\newline3C4)O2)O)C=C
& -9.2\\
\hline
CFW-008 & [C@]12(C[C@@H]1CCCC[C@H]1C\newline CN(C(=O)N1NC(=O)C[C@@H](NC(\newline=O)N2)Cc1cccc(c1)NC(=O)C1=CN(N\newline C1)CC1CC1)c1ccc2c(c1)OCO2)C(\newline=O)NN
& -9.2\\
\hline
CFW-009 & c1(ccc2c(c1)N/C(=C/1$\backslash$CC=CC=C1)/\newline N1C3=C1[C@@H]1CCC[C@H]2C\newline N2CCN(CC(=O)[C@H]([C@@H](C3)\newline O1)NNO)C(=O)N2)OC & -9.1\\
\hline
CFW-010 & C$\backslash$1(=C/C2=C(CCc3ccccc3N3C=C/C(=\newline C/CC/C=C/O)/C(=O)[C@]3(n3cccc\newline13)O)SC[C@@H]1CO[C@H](NN21\newline)c1ccccc1)/CC(=O)O
& -9.0\\
\hline
CFW-011 & [C@H]1(NC(=O)C/C(=C$\backslash$NC=C2CC2\newline)/C=C(/COc2ccccc2[C@H](NC(=O)C[\newline C@H](NC1=O)CC1CCCCC1)CO)$\backslash$C=\newline C$\backslash$N1NCNC2=C1CNC2)C(C)C
& -9.0\\
\hline
CFW-012 & [C@@H]1(CC[C@H]2C[C@H](CC(=\newline C)C2)[C@H]2CN3CCN(C[C@@H]3C\newline(=O)[C@H]3[C@H]1CCC=C3C2(C)C)\newline Cc1cccc(c1)OC)C=C
& -8.9\\
\hline

\end{tabular}
\end{center}
\newpage

\begin{center}
\begin{tabular}{|p{0.33\columnwidth}|p{0.33\columnwidth}|p{0.33\columnwidth}|}
\hline
Name of the drug& SMILES& Binding Affinity\\
\hline

Remdesivir & CCC(CC)COC(=O)C(C)NP(=O)(OCC1\newline C(C(C(O1)(CN)C2=CC=C3N2N=CN=\newline C3N)O)O)OC4=CC=CC=C4
& -7.3\\
\hline
Ribavirin & CCC(CC)COC(=O)C(C)NP(=O)(OCC1\newline C(C(C(O1)(CN)C2=CC=C3N2N=CN=\newline C3N)O)O)OC4=CC=CC=C4
& -6.0\\
\hline
Arbidol/Umifenovir & CCOC(=O)C1=C(N(C2=CC(=C(C(=C2\newline
1)CN(C)C)O)Br)C)CSC3=CC=CC=C3 & -6.1\\
\hline
Favipiravir & C1=C(N=C(C(=O)N1)C(=O)N)F & -5.4\\
\hline
Lopinavir(isometric) & CC1=C(C(=CC=C1)C)OCC(=O)N[C@\newline
@H](CC2=CC=CC=C2)[C@H](C[C@H\newline
](CC3=CC=CC=C3)NC(=O)[C@H](C\newline
(C)C)N4CCCNC4=O)O & -8.0\\
\hline
Dexamethasone & CC1CC2C3CCC4=CC(=O)C=CC4(C3\newline
(C(CC2(C1(C(=O)CO)O)C)O)F)C & -6.8\\
\hline

\end{tabular}
\end{center}

\begin{figure}[htbp]
\centering 
\includegraphics[width=1.0\textwidth]{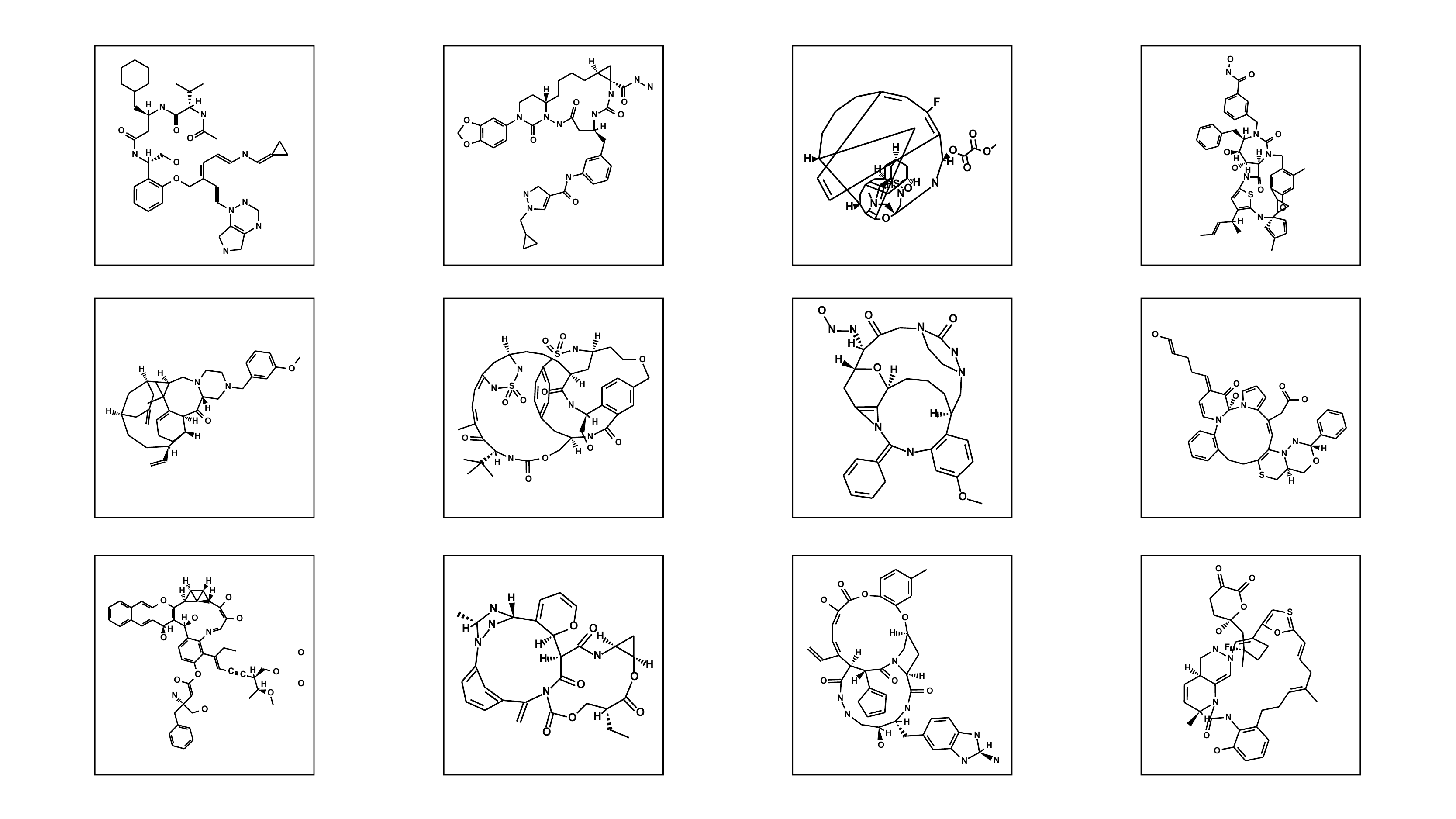}
\caption{The 2D Structure of our generated molecules} 
\label{Fig.main2} 
\end{figure}
\begin{figure}[htbp]
\centering 
\includegraphics[width=1.0\textwidth]{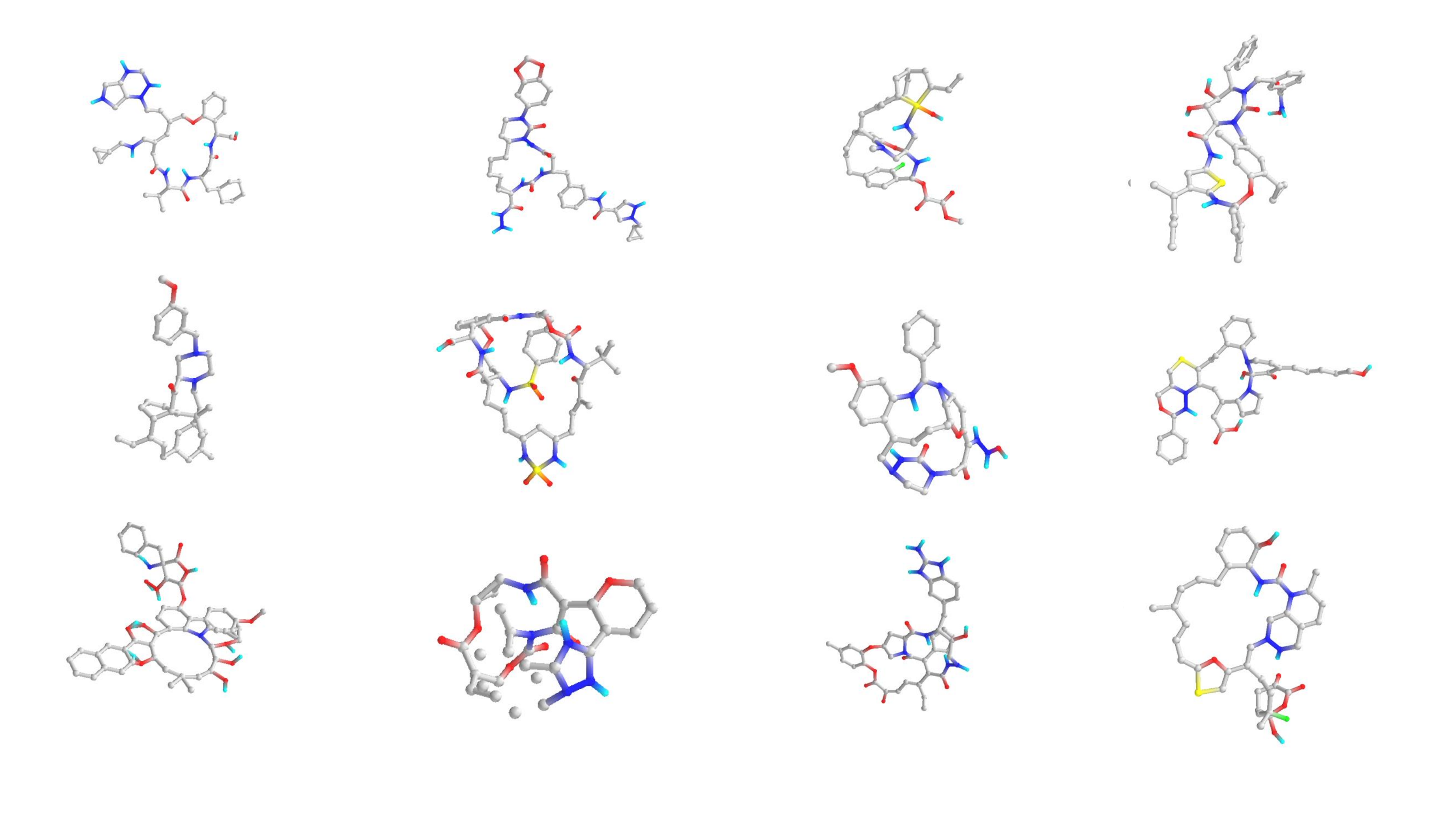}
\caption{The Balls-and-Sticks Structure of our generated molecules} 
\label{Fig.main2} 
\end{figure}
\newpage
\subsection{SwissADME Check}
SwissADME \cite{swissadme} is a website that enables the computational prediction of ADME properties, physiochemical descriptors, druglike nature and other parameters related to drug discovery. It is one of the most well-regarded software in the field of pharmacokinetics, and the research paper describing it has more than one thousand citations according to Microsoft Academic. In the context of our study, four parameters are of the greatest concern (in order of importance): the Brenk structural alert, Synthetic Accessibility, Bioavailability Score, and the PAINS alert. All other parameters, such as lipophilicity and water solubility, are less relevant in that they are covered by the above five or do not apply well to our problem.

The Brenk structural alert is a list of 105 molecule fragments identified to be putatively toxic, chemically reactive, metabolically unstable, or has other undesirable qualities by Brenk R. et al \cite{adme1}. This parameter is important because molecules that set off this alert has a higher risk of being toxic or in other ways unsuitable for drug use.

Synthetic accessibility (SA) is the estimated difficulty to generate a molecule in laboratory conditions. The lower the SA score, according to estimation, the easier it is to create this molecule, and the lower the costs of manufacturing drugs based on it.

The Bioavailability score, or Abbtt bioavailability, predicts the percentage of drug molecule that might reach the systemic circulation \cite{adme2}. More specifically, it is the probability that a compound will have above 10\% oral bioavailability in the rat. This metric matters because drugs with high bioavailability can achieve greater results with smaller doses, this influencing manufacturing costs.

The PAINS (pan assay interference compounds) alert, developed by Baell et al \cite{adme3}.,  indicates the presence of PAINS functional groups, such as  toxoflavin, isothiazolones, hydroxyphenyl hydrazones, and curcumin \cite{adme4}– parts of molecules that react nonspecifically with numerous targets instead of only the desired one, and frequently yield a false positive output.

\begin{center}
\begin{tabular}{|p{0.15\columnwidth}|p{0.15\columnwidth}|p{0.15\columnwidth}|p{0.15\columnwidth}|p{0.15\columnwidth}|p{0.15\columnwidth}|}
\hline
Name of the drug& Binding Affinity
& Brenk Structual Alert
& Synthetic Assessbility
& Bioavialability Score
 & PAINS
\\
\hline
CFW-001 & -11.9 & 2 alerts & 9.84 & 0.17 & 0 alerts\\
\hline
CFW-002 & -9.9 & 4 alerts & 8.61 & 0.17 & 0 alerts\\
\hline
CFW-003 & -9.7 & 2 alerts & 7.61 & 0.17 & 0 alerts\\
\hline
CFW-004 & -9.6 & 3 alerts & 9.43 & 0.17 & 0 alerts\\
\hline
CFW-005 & -9.4 & 1 alerts & 8.4 & 0.17 & 0 alerts\\
\hline
CFW-006 & -9.3 & 0 alerts & 10 & 0.17 & 0 alerts\\
\hline
CFW-007 & -9.2 & 3 alerts & 9.35 & 0.55 & 0 alerts\\
\hline
CFW-008 & -9.2 & 2 alerts & 7.34 & 0.17 & 0 alerts\\
\hline
CFW-009 & -9.1 & 1 alerts & 8 & 0.17 & 0 alerts\\
\hline
CFW-010 & -9.0 & 2 alerts & 7.11 & 0.17 & 0 alerts\\
\hline
CFW-011 & -9.0 & 0 alerts & 7.91 & 0.17 & 0 alerts\\
\hline
CFW-012 & -8.9 & 1 alerts & 6.79 & 0.17 & 0 alerts\\
\hline
Remdesivir & -7.3 & 1 alerts & 6.33 & 0.17 & 0 alerts\\
\hline
Ribavirin & -6.0 & 0 alerts & 3.89 & 0.55 & 0 alerts\\
\hline
Favipiravir & -5.4 & 0 alerts & 2.08 & 0.55 & 0 alerts\\
\hline
Lopinavir\newline(isometric)
 & -8.0 & 0 alerts & 5.67 & 0.55 & 0 alerts\\
\hline
Arbidol/\newline
Umifenovir
 & -6.1 & 0 alerts & 3.57 & 0.55 & 1 alerts\\
\hline
Dexamethasone
 & -6.8 & 0 alerts & 5.47 & 0.55 & 0 alerts\\
\hline
\end{tabular}
\end{center}

The table above is the results of SwissADME check. The analysis below provides detail explanation of each parameters.

SwissADME provides some valuable parameters that affect the quality of a drug molecule, whereas these parameters are not the primary focus of our study. While a drug that performs well in these parameters doubtless has its merits, and a drug that performs poorly has its drawbacks, we nevertheless deem binding affinity with the virus – the core factor that represents the upper limit of the drug’s effectiveness. Due to these reasons, we do not directly incorporate ADME variables in our GCGVAE model, but instead merely add a size penalty to our GA (because large molecules generally have poor ADME performance).
\section{Conclusions}
According to our docking results with 6LU7, the crystal structure of SARS-CoV-2 main protease in complex with an inhibitor N3 \cite{6lu7}, the binding affinity of Remedsivir, Favipiravir, Arbidol \cite{remdesivir,other_drug}, and several widely tested and produced drugs that target to cure SARS-CoV-2 with 6LU7, is significantly less than the binding affinity of the drug that we generated using our model with 6LU7. 
\\
Our model of generating drug molecules comes with less cost and is more reliable compared to traditional methods. Because our results can be easily stored and tested on ordinary computers using particular software, the cost of testing whether the drug molecules are effective is comparatively low. Also, because virtual screening is more stable than the traditional way of testing drugs since docking ligands of drug molecules using computers cannot be interfered by confounding variable that may take place in traditional labs, it guarantees the reliability of our result. 
\\
Also, our model is not restricted to generating drugs inhibiting SARS-CoV-2. Unlike traditional methods of discovering drugs, which usually requires a specific and detailed analysis on the structure of the virus, the model we used is not restricted to a specific virus, which, in this case, is SARS-Cov-2. GCGVAE enables us to generate drugs correspondent to inhibit any virus if the structure of the main protease of the virus is entered into the model, which gives us high flexibility and capability in generating drugs targeted to cure viruses other than SARS-CoV-2 in the future.
\\
Another unique characteristic of our model is that it increases the variability of ordinary CGVAE. While employing CGVAE alone does not have high flexibility and novelty in generating output, which limits the output to a relatively small range, we employ a GA after the process of CGVAE by using an innovative method of crossover and mutation. This action enables our model to generate a more unique set of output than traditional CGVAE models and thus increase the number of drug structures to be tested. 

 \bibliographystyle{unsrt}  
 \bibliography{references.bib}  






\end{document}